\documentclass{article}

\PassOptionsToPackage{numbers, compress}{natbib}
\usepackage[preprint]{neurips_2021}

\usepackage[utf8]{inputenc} 
\usepackage[T1]{fontenc}    
\usepackage{hyperref}       
\usepackage{url}            
\usepackage{booktabs}       
\usepackage{amsfonts}       
\usepackage{nicefrac}       
\usepackage{microtype}      
\usepackage{xcolor}         

\title{Compositionality-Aware Graph2Seq Learning}

\author{%
  Takeshi D. Itoh \\
  Graduate School of Science and Technology \\
  Nara Institute of Science and Technology \\
  \texttt{itoh.takeshi.ik4@is.naist.jp} \\
  \And
  Takatomi Kubo \\
  Graduate School of Science and Technology \\
  Nara Institute of Science and Technology \\
  \texttt{takatomi-k@is.naist.jp} \\
  \And
  Kazushi Ikeda \\
  Graduate School of Science and Technology \\
  Nara Institute of Science and Technology \\
  \texttt{kazushi@is.naist.jp}
}

\usepackage{amsmath}
\usepackage{bm}
\usepackage{graphicx}
\usepackage{multirow}

\newcommand{\Real}{{\mathbb{R}}}

\newcommand{\EG}{\textit{e}.\textit{g}., }

\newcommand{\SecLab}[1]{\label{sec:#1}}
\newcommand{\SecRef}[1]{Section~\ref{sec:#1}}

\newcommand{\FigLab}[1]{\label{fig:#1}}
\newcommand{\FigRef}[1]{Figure~\ref{fig:#1}}

\newcommand{\TblLab}[1]{\label{tbl:#1}}
\newcommand{\TblRef}[1]{Table~\ref{tbl:#1}}

\newcommand{\EqLab}[1]{\label{eq:#1}}

\newcommand{\MlapGraph}{{G}}
\newcommand{\MlapGraphEmb}[1]{{\bm{h}_{#1}}}
\newcommand{\MlapLayerEmb}[2]{{\bm{h}_{#1}^{(#2)}}}

\newcommand{\MlapNodes}{{\mathcal{N}}}
\newcommand{\MlapNode}{{n}}
\newcommand{\MlapNodePrime}{{n'}}

\newcommand{\MlapNodeEmb}[2]{{\bm{h}_{#1}^{({#2})}}}

\newcommand{\MlapNumLayer}{{L}}
\newcommand{\MlapLayer}{{l}}

\newcommand{\MlapAggFn}{{f_\mathrm{agg}}}

\newcommand{\MlapLayerGateFn}[1]{{f_\mathrm{gate}^{(#1)}}}

\newcommand{\MlapEmbDim}{{d}}

\newcommand{\LstmState}[1]{{\bm{h}^\mathrm{dec}_{#1}}}
\newcommand{\LstmMemory}[1]{{\bm{m}^\mathrm{dec}_{#1}}}
\newcommand{\LstmContext}[1]{{\bm{c}^\mathrm{dec}_{#1}}}
\newcommand{\LstmStep}{{t}}
\newcommand{\LstmMaxStep}{{T}}
\newcommand{\LstmIn}[1]{{\bm{x}^\mathrm{dec}_{#1}}}
\newcommand{\LstmOut}[1]{{\bm{y}^\mathrm{dec}_{#1}}}
\newcommand{\LstmVoc}{{\bm{E}^\mathrm{voc}}}
\newcommand{\LstmVocWord}[1]{{\bm{E}^\mathrm{voc}_{#1}}}
\newcommand{\LstmNumVoc}{{N^\mathrm{voc}}}

\newcommand{\LinearPos}{{i}}
\newcommand{\LinearVoc}{{\bm{E}^\mathrm{voc}_\LinearPos}}
\newcommand{\LinearVocWord}[1]{{\bm{E}^\mathrm{voc}_{\LinearPos,#1}}}
\newcommand{\LinearMaxPos}{{I}}

\begin{document}

\maketitle

\begin{abstract}
    Graphs are a highly expressive data structure, but it is often difficult for humans to find patterns from a complex graph.
Hence, generating human-interpretable sequences from graphs have gained interest, called \emph{graph2seq} learning.
It is expected that the compositionality in a graph can be associated to the compositionality in the output sequence in many graph2seq tasks.
Therefore, applying compositionality-aware GNN architecture would improve the model performance.
In this study, we adopt the multi-level attention pooling (MLAP) architecture, that can aggregate graph representations from multiple levels of information localities.
As a real-world example, we take up the extreme source code summarization task, where a model estimate the name of a program function from its source code.
We demonstrate that the model having the MLAP architecture outperform the previous state-of-the-art model with more than seven times fewer parameters than it\footnote{The code is provided in our github repository: \protect\url{https://github.com/iTakeshi/mlap-graph2seq}.}.

\end{abstract}

\section{Introduction\SecLab{introduction}}

Graphs are a type of data structure that is capable of encoding a set of rich pairwise information among elements.
However, in exchange for the expressivity of a complex graph, it is often difficult for humans to find patterns in such a graph.
Therefore, there exists a growing interest in learning to generate a human-interpretable output from a graph.
In this study, we focus on generating a sequence of discrete symbols from a graph, known as \emph{graph2seq} tasks.
For example, if we could generate a natural question from knowledge graphs, we could use them to develop an artificial intelligent agent that can interact with people~\citep{sakaguchi2013discriminative,seyler2017knowledge}.
Furthermore, we could solve a path-planning problem as an instance of graph2seq tasks when we regard the input map as a graph~\citep{hu2004knowledge}.

Moreover, traditional sequence to sequence (\emph{seq2seq}) tasks, like neural machine translation (NMT), could benefit from transforming them into graph2seq tasks.
Seq2seq studies have achieved significant successes by introducing the encoder-decoder architecture.
Pioneering studies applied recurrent neural networks to encode the input sequence~\citep{cho2014learning,sutskever2014sequence}, and later transformer-based architectures dominated the field~\citep{vaswani2017attention,devlin2018bert}.
Either way, the input to the models are processed as one-dimensional \emph{sequences}.
On the other hand, many kinds of input sequences in seq2seq tasks can actually be represented as \emph{graphs} with enriched information.
In many cases, we can derive rule-based conversions from raw input sequences to graphs by considering the context and the compositionality in the sequence---\EG we can convert a natural language sentence to a graph by analyzing its grammatical structure and dependency relations.
Because of such additional information, we can expect that graph2seq models utilizing graph structures in seq2seq tasks have a potential to learn improved representations~\citep{fernandes2019structured}.

Therefore, developing a graph2seq model using graph neural networks (GNNs) has recently piqued the interest.
For example, \citet{marcheggiani2018deep} introduced a simple graph2seq model that combines a GCN encoder and an LSTM decoder to generate a text description from a resource description framework (RDF) graph or a semantic dependency graph.
\citet{xu2018graph2seq} proposed an attention-based graph2seq model for graph reasoning tasks like bAbI or shortest path search.
\citet{chen2020toward} and \citet{wei2021graph} applied similar attentional graph2seq techniques in question generation and opinionated text summarization tasks, respectively.
\citet{zhu2019modeling} and \citet{cai2020graph} proposed graph-transformer-based graph2seq methods, which consider the relationship not only between neighbor nodes but also between distant nodes.
However, these methods did not consider the compositionality and hierarchy of the relations among nodes.
\citet{chen2020toward} proposed the Heterogeneous Graph Transformer, that explicitly considers the hierarchy in the input graphs.
It splits the original graph into multiple subgraphs and computes the graph representation by aggregating the subgraph representations.
It improved the performances of multiple text generation tasks including NMT, owing to its ability to utilize the compositional nature of graphs.
However, HetGT only considers single-level hierarchy.

Our assumption in this study is that considering multi-level compositionality in the graphs is effective to learn good representation for generating sequences.
The output sequences of graph2seq tasks often have multi-level compositional nature.
For example, words in a sentence form local phrases---\EG an adjective generally modifies a noun next to the adjective---and then phrases link each other and form longer parts.
Here, we speculate that the compositionality in a graph can be associated to the compositionality of the output sequence.

Based on the assumption, we apply the multi-level attention pooling architecture (MLAP)~\citep{itoh2022multi} to graph2seq tasks.
Because of the message passing procedure~\citep{gilmer2017neural} propagating node information between direct neighbors, information locality of node embeddings are small in lower layers and it expands toward higher layers.
MLAP is a GNN architecture that computes layer-wise graph representations using an attention pooling operation in each hidden GNN layers, and then aggregate them as the final graph representation.
Therefore, GNN models with MLAP architecture can utilize the graph information in multiple levels of localities, achieving a compositionality-aware graph representation learning.

In this study, we aim to demonstrate that the MLAP architecture is effective for a graph2seq learning.
As a real-world example, we take up the ``extreme source code summarization'' task---generating a short natural language summary of a software program snippet~\citep{allamanis2016convolutional}.
Although program source code is usually written in a programming language (\EG C, Java, Python) and represented as a sequence of tokens, it can be converted into graphs called abstract syntax trees (ASTs) by analyzing the syntactic structure.
Also, each function in source code has a name summarizing the behavior of the part of program.
Therefore, a task to estimate the function name from a program graph is a good example of graph2seq learning.
There has already been multiple attempts to apply graph2seq-style model to this task.
For example, \citet{allamanis2018learning} and \citet{fernandes2019structured} trained gated graph neural network~\citep{li2016gated} on ASTs.
However, many of the methods proposed in those studies do not explicitly consider the compositionality in source code.
Program graphs have both short and long dependencies in them, that is, short dependencies form individual operations and longer ones structure complex algorithms.
Therefore, considering compositionality in the program graphs can improve the performance.
In this study, we train a MLAP model on a source code dataset and show that our proposed method outperforms the previous state-of-the-art (SoTA) model\footnote{\protect\url{https://ogb.stanford.edu/docs/leader\_graphprop/\#ogbg-code2}, accessed November 21, 2021. \textit{Note that the SoTA record we refer in this technical report was broken on January 15, 2022 (just before we submit this paper to arXiv).}}.

The rest of this paper is organized as follows:
\SecRef{methods} extends the MLAP framework for graph2seq learning,
\SecRef{experiments} describes the the source code summarization experiments and results,
and \SecRef{results} and \SecRef{discussion} demonstrates and discusses the results.

\section{MLAP for graph2seq Learning\SecLab{methods}}

In this section, we extend the MLAP architecture introduced in \citet{itoh2022multi} for graph2seq learning.
We first briefly summarize the formulation of MLAP as a graph encoder and then introduce sequence decoders on top of the encoder.

\subsection{Graph Encoder}

Suppose we are given a graph $\MlapGraph$ composed of a set of nodes $\MlapNodes$ and an $\MlapNumLayer$-layer GNN model taking the graph as its input.
Let $\MlapNodeEmb{\MlapNode}{\MlapLayer}$ denote the vector embedding of a node $\MlapNode \in \MlapNodes$ in the $\MlapLayer$-th ($1 \le \MlapLayer \le \MlapNumLayer$) layer.
For both decoders, suppose we have computed the layer-wise representations $\MlapLayerEmb{\MlapGraph}{\MlapLayer}$ and the final graph representation $\MlapGraphEmb{\MlapGraph}$ as shown in \citet{itoh2022multi}, that is
\begin{align}
    \MlapLayerEmb{\MlapGraph}{\MlapLayer} &= \sum_{\MlapNode \in \MlapNodes} \mathrm{softmax} \left( \MlapLayerGateFn{\MlapLayer} (\MlapNodeEmb{\MlapNode}{\MlapLayer}) \right) \MlapNodeEmb{\MlapNode}{\MlapLayer} \EqLab{mlap-hl} \quad \forall \ \MlapLayer \in \{1, \dots, \MlapNumLayer\}, \\
    \MlapGraphEmb{\MlapGraph} &= \MlapAggFn \left( \left\{ \MlapLayerEmb{\MlapGraph}{\MlapLayer} \ \middle| \ \MlapLayer \in \{1, \dots, \MlapNumLayer\} \right\} \right),
\end{align}
where $\MlapLayerGateFn{\MlapLayer}$ is a function used to compute the attention score, for which a two-layer neural network was used in this study, and $\MlapAggFn$ is an aggregation function.
Readers are referred to the original MLAP paper~\citep{itoh2022multi} for the detailed formulation.

\subsection{Sequence Decoder}

We evaluate two types of sequence decoders in this study: \emph{Linear} and \emph{LSTM}.
The former selects elements in a sequence independently, whereas the latter considers the context through the sequence.

\subsubsection{Linear Decoder}

The most simple implementation of a decoder uses a linear word classifier solely based on the final graph representation, independently for each position in the decoded sequence.
That is, we use five independent classifiers if we want to decode five-word sequences.

The model learns \emph{position-wise} embedded representation matrix for the target vocabulary $\LinearVoc \in \Real^{\LstmNumVoc\times\MlapEmbDim}$ for each position $\LinearPos$, where $\LstmNumVoc$ is the size of vocabulary that consists the generated sequence.
Here, the words in the vocabulary with ids 0, 1, and 2 have special meanings: the start of a sentence (SOS), an unknown word (UNK), and the end of a sentence (EOS).

The selection probability for a word $w$ from vocabulary in each position is computed as
\begin{equation}
    p_{\LinearPos,w} = \mathrm{softmax} \left( \MlapGraphEmb{\MlapGraph} \cdot \LinearVocWord{w} + b_{\LinearPos,w} \right) \quad \forall \ \LinearPos \in \{1,\dots,\LinearMaxPos\},
\end{equation}
where $b_\LinearPos$ is a bias parameter and $\LinearMaxPos$ is the length of decoded sequences.
If one or more classifiers select the word EOS for any positions, the words after the first EOS are ignored.

\subsubsection{LSTM Decoder}

The aforementioned Linear decoder is rather naive, and each word is decoded independent of other words.
Thus, it cannot consider the context among decoded words.
Instead, we can use a LSTM-based decoder with an attention mechanism~\citep{luong2015effective} to generate a sequence from those representations.

In this case, the model learns an \emph{position-invariant} embedded representation matrix for the target vocabulary $\LstmVoc \in \Real^{\LstmNumVoc\times\MlapEmbDim}$.

We first initialize the LSTM state $\LstmState{0}$ and the memory cell $\LstmMemory{0}$ using the aggregated graph representation as
\begin{align}
    \LstmState{0}  &= \MlapGraphEmb{\MlapGraph}, \\
    \LstmMemory{0} &= \MlapGraphEmb{\MlapGraph}.
\end{align}
At the step $\LstmStep \ (\LstmStep = 1,\dots,\LstmMaxStep)$, the state and the memory cell are updated by a standard LSTM~\citep{sak2014long}.
\begin{align}
    \LstmState{\LstmStep}, \LstmMemory{\LstmStep} &= \mathrm{LSTM}(\LstmIn{\LstmStep}, \LstmState{\LstmStep - 1}, \LstmMemory{\LstmStep - 1}), \\
    \LstmIn{\LstmStep} &= \begin{cases}
        \LstmVocWord{\mathrm{SOS}} \quad & \text{if $\LstmStep = 0$}, \\
        \LstmOut{\LstmStep - 1}          & \text{otherwise}.
    \end{cases}
\end{align}
Here, $\LstmOut{\LstmStep}$ is the output of the decoder at $\LstmStep$ (see following).

To determine the output, the model first compute a context vector using an attention mechanism over the layer-wise and final graph representations.
Here, for notation simplicity, we consider $\MlapLayerEmb{\MlapGraph}{\MlapNumLayer+1}$ be $\MlapGraphEmb{\MlapGraph}$, and then the computation is as
\begin{equation}
    \LstmContext{\LstmStep} = \sum_{\MlapLayer=1}^{\MlapNumLayer+1} \mathrm{softmax}\left(\MlapLayerEmb{\MlapGraph}{\MlapLayer} \cdot \LstmState{\LstmStep}\right) \MlapLayerEmb{\MlapGraph}{\MlapLayer}.
\end{equation}
Then, it computes the output as follows.
\begin{equation}
    \LstmOut{\LstmStep} = \tanh \left( \mathrm{LayerNorm} \left( \bm{W}^\mathrm{dec} \cdot [\LstmContext{\LstmStep}; \LstmState{\LstmStep}] + \bm{b}^\mathrm{dec}\right) \right),
\end{equation}
where $\bm{W}^\mathrm{dec}$ and $\bm{b}^\mathrm{dec}$ are trainable parameters.
The selection probability for each word in the vocabulary at $\LstmStep$ is computed as
\begin{equation}
    p_{\LstmStep,w} = \mathrm{softmax} \left( \LstmOut{\LstmStep} \cdot \LstmVocWord{w} + b_w \right),
\end{equation}
where $b_w$ is a bias parameter.
The LSTM decoder stops when the word EOS is generated or the step $\LstmStep$ reaches the maximum step $\LstmMaxStep$.

\section{Experiments\SecLab{experiments}}


\subsection{Task and Dataset}

To evaluate the MLAP-graph2seq model, we employed the ``extreme source code summarization'' task~\citep{allamanis2016convolutional}, wherein a model predicted the function name for given source code snippet (\FigRef{seq-task}).
Because a function is commonly named in such a way that the name describes the behavior of the function, recovering the function name from its body would be one of the fundamental tasks that evaluates the models' capability to understand source code semantics.

Here, we used the ogbg-code2 dataset from OGB collection~\citep{hu2020open}.
It contains 452,741 ASTs, each of which contains 125.2 nodes and 124.2 edges on average, extracted from 13,587 OSS projects.
The average length of the ground-truth sequence is 2.25.
Each node has a 3-dimensional discrete-valued feature containing the depth of node counted from the root of the AST, the type of AST node (\EG \texttt{If} or \texttt{Num}), and the node attribute.
Here, only the leaf nodes have node attributes, usually the token string associated to the node.
Non-leaf nodes have blank attributes.

\begin{figure*}[t]
    \centering
    \includegraphics[width=0.9\textwidth]{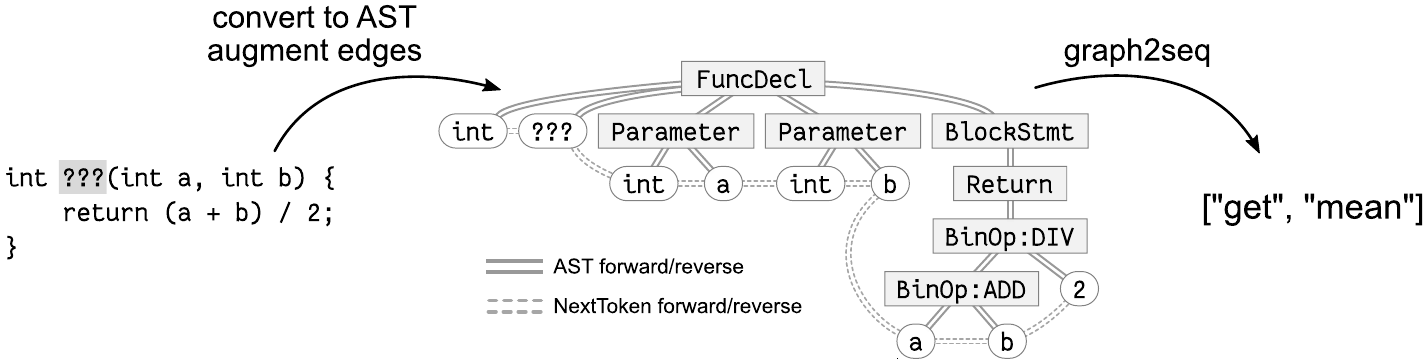}
    \caption[Extreme code summarization task]{
        A schematic illustration of the extreme code summarization task.
        A source code snippet with a masked function name, is provided as the input.
        The snippet is converted into an AST, and then augmented with additional types of edges in preprocess.
        Models are trained to recover the original function name (\texttt{get\_mean}).
    }
    \FigLab{seq-task}
\end{figure*}


\subsection{Preprocess}

As introduced in \citet{allamanis2018learning}, we augmented the AST graph with additional types of edges.
The original graphs in the ogbg-code2 dataset only have one type of edges that point from parent nodes to child nodes, and thus models suffer from low efficiency in information diffusion during message passing.
Therefore, we added three additional types of edges to the graphs (\FigRef{seq-task}):
\begin{enumerate}
    \item Edges pointing from child nodes to parent nodes (reverse edges of original AST edges).
    \item Edges pointing from preceding tokens to following tokens (NextToken edges~\citep{allamanis2018learning}; only among leaf nodes).
    \item Edges pointing from following tokens to preceding tokens (reverse NextToken edges).
\end{enumerate}


\subsection{Model Configuration}

We used MLAP-Weighted architecture throughout the experiment, and evaluated both Linear and LSTM decoders.
We set the maximum length of the decoded sequence ($\LinearMaxPos$ or $\LstmMaxStep$) as 5.

We used graph isomorphism network (GIN)~\citep{xu2019how} as the message passing layer.
For each decoder type (Linear or LSTM), we selected the best model configuration among these conditions: number of layers (5 or 6), residual connection (+ or $-$), and GraphNorm~\citep{cai2020graphnorm} (+ or $-$).
If a model has residual connection, the node presentation before message passing is added to the node representation after dropout, as in
\begin{equation*}
    \MlapNodeEmb{\MlapNode}{\MlapLayer} = \mathrm{DropOut}\left(\mathrm{ReLU}\left(\mathrm{GraphNorm}\left(\mathrm{GIN}\left(\left\{\MlapNodeEmb{\MlapNodePrime}{\MlapLayer-1} \middle| \MlapNodePrime \in \MlapNodes \right\}\right)\right)\right)\right) + \MlapNodeEmb{\MlapNode}{\MlapLayer-1}.
\end{equation*}

We optimized the model using the Adam optimizer~\citep{kingma2015adam}.
The models were trained for 50 epochs against a cross-entropy loss computed for each word in the output sequence.
The initial learning rate was set to $5\times10^{-4}$ and decayed by $\times 0.2$ after 3 epochs without improvement in the validation F1 score.
The batch size was 256.


\subsection{Performance Evaluation}

As proposed by \citet{alon2019code2vec}, we used an F1 score between predicted words and ground-truth words to evaluate the model performance.
Given the unique and order-agnostic sets of predicted words $S_p$ and ground-truth words $S_t$, we defined the metric as
\begin{align*}
    \mathrm{TP} &= |S_p \cap S_t|, &
    \mathrm{FP} &= |S_p \setminus S_t|, &
    \mathrm{FN} &= |S_t \setminus S_p|, \\
    \mathrm{precision} &= \frac{\mathrm{TP}}{\mathrm{TP} + \mathrm{FP}}, &
    \mathrm{recall} &= \frac{\mathrm{TP}}{\mathrm{TP} + \mathrm{FN}}, &
    \mathrm{F1} &= \frac{2 \times \mathrm{precision} \times \mathrm{recall}}{\mathrm{precision} + \mathrm{recall}}.
\end{align*}
We calculated an F1 score for each graph and calculated average over the dataset.
We trained 10 models with different random seeds, and evaluate the performance by the mean and the unbiased standard deviation.

As baseline models, we compared the performance of our model to the naive GIN architecture without MLAP, code2seq~\citep{alon2019code2seq}, and Graph Multi-head Attention Neural Network (GMAN)~\citep{zhang2021gman}, which is the previous SoTA model in the OGB's leaderboard to date.
We trained the naive GIN and code2seq models ourselves because there are no available pre-trained models.
Note that the definition of the F1 score used in original code2seq code is different from OGB (code2seq first sums up TP, FP, and FN over the dataset and then computes precision, accuracy, and F1), and we replaced the definition with that of OGB.
Also, we trained the code2vec models for 100 epochs because of slow convergence.
For GMAN, we referred to the performance on the OGB's leaderboard.

\section{Results\SecLab{results}}

\begin{table}[t]
    \caption{Summary of the model performance (\textit{mean} $\pm$ \textit{unbiased standard deviation}) on ogbg-code2.}
    \resizebox{\textwidth}{!}{
    \begin{tabular}{lll|ccc}
        Decoder                 & Model         & Configuration            & \#Params & Validation F1       & Test F1                      \\ \hline\hline
        \multirow{2}{*}{Linear} & MLAP-Weighted & $L=6$, Res (+), GN ($-$) & 8.6M     & 0.1649 $\pm$ 0.0014 & \textbf{0.1792} $\pm$ 0.0017 \\
                                & naive         & $L=6$, Res (+), GN ($-$) & 8.3M     & 0.1622 $\pm$ 0.0018 & 0.1768 $\pm$ 0.0019          \\ \hline
        \multirow{2}{*}{LSTM}   & MLAP-Weighted & $L=5$, Res (+), GN (+)   & 5.0M     & 0.1602 $\pm$ 0.0021 & 0.1762 $\pm$ 0.0037          \\
                                & naive         & $L=6$, Res (+), GN (+)   & 4.6M     & 0.1596 $\pm$ 0.0024 & 0.1744 $\pm$ 0.0030          \\ \hline\hline
        \multicolumn{3}{c|}{GMAN~\citep{zhang2021gman}}                    & 63.7M    & 0.1631 $\pm$ 0.0090 & 0.1770 $\pm$ 0.0012          \\
        \multicolumn{3}{c|}{code2seq~\citep{alon2019code2seq}}             & 26.7M    & 0.1495 $\pm$ 0.0010 & 0.1549 $\pm$ 0.0010
    \end{tabular}
    }
    \TblLab{seq-results}
\end{table}

\begin{table}[t]
    \centering
    \caption{The statistical analysis results. E.S.: effect size (Cohen's $d$).}
    \small
    \begin{tabular}{l|cc}
        Comparison                                                           & $p$         & E.S. \\ \hline\hline
        MLAP-Weighted-Linear \textit{vs}.\ naive-Linear                      & *0.007      & 1.37 \\
        MLAP-Weighted-Linear \textit{vs}.\ GMAN~\citep{zhang2021gman}        & *0.003      & 1.53 \\
        MLAP-Weighted-Linear \textit{vs}.\ code2seq~\citep{alon2019code2seq} & *$<10^{-5}$ & 17.7
    \end{tabular}
    \TblLab{seq-stat}
\end{table}

\TblRef{seq-results} summarizes the model performance on the ogbg-code2 dataset.
We selected the best models according to the validation performance.
The MLAP-Weighted model with the Linear decoder achieved the test F1 score of 0.1792 $\pm$ 0.0016.
It outperformed the previous SoTA GNN model (GMAN; 0.1770 $\pm$ 0.0012) or non-GNN model (code2seq; 0.1549 $\pm$ 0.0009).

For the tested ogbg-code2 dataset, the LSTM decoder did not perform as good as the Linear decoder.
Nevertheless, the MLAP-Weighted model outperformed the naive architecture.

We statistically compared the performance of the MLAP-Weighted model with the Linear decoder to other models (\TblRef{seq-stat}).
As a result, it was revealed that our proposed model performed significantly better than the baseline models.
Also, the effect size evaluated by Cohen's $d$~\citep{cohen1988statistical} demonstrated the performance difference between our model and others were large.

\section{Discussion\SecLab{discussion}}

In this study, we extended the MLAP architecture for graph2seq tasks.
Our proposed model outperformed the naive architecture and GMAN in the extreme source code summarization task and outperformed the previous SoTA model.
It indicates that explicitly using the compositional graph information is beneficial in learning to generate sequences from graphs.
Here, we emphasize that our best model using MLAP has only 8.6 million parameters, which is less than a seventh of the previous SoTA model.
It further supports our hypothesis that utilizing the compositionality helps the model to capture the structural information in graphs.

Code2seq, which is one of the best non-GNN program comprehension model to the best of our knowledge, performed worse than both our model and GMAN, even though code2seq is designed to exploit the domain knowledge in software engineering as much as possible.
We could consider the ``domain knowledge'' used in code2seq or other ML-based program comprehension studies as a set of human experiences in which human programmers mentally construct and interpret the program graph like ASTs.
Therefore, it may be enough to use GNNs, which are highly capable of capturing the graph structure, rather than building program comprehension-specific model full of domain knowledge.

Under the current experimental settings, the Linear decoder consistently performed better than the LSTM decoder.
We can consider several hypotheses that explain the reason of the results.
First, our hyperparameter tuning for the LSTM decoder might not be sufficient.
As observed in \TblRef{seq-results}, the number of parameters of the model using LSTM decoder is smaller than that with Linear decoder.
This would limit the expressivity of the LSTM decoder, and we should be able to further improve the performance in this regard (\EG adopting bidirectional LSTM decoder~\citep{alsabahi2018bidirectional}).
Second, the comparably complex LSTM decoder is harder to train with a medium-sized ogbg-code2 dataset.
It only contains approximately $4.5 \times 10^5$ data points, with each ground-truth sequence having only 2.25 words, which may not be enough for LSTM to capture the context in the output sequences.
By contrast, the LSTM decoder should be better than the Linear decoder for tasks that require the models to generate longer sequences.

To further improve the performance of MLAP-graph2seq models, seeking for various decoder implementation would be needed.
For example, instead of uni-directional LSTM used in this study, we may use bi-directional LSTM as the decoder~\citep{alsabahi2018bidirectional}, which might be beneficial in capturing the context of the output sequences.
Furthermore, using transformer-like architecture~\citep{vaswani2017attention} can be another option.

\subsection{Concluding Remarks}
In this paper, we propose a graph2seq model using the MLAP architecture.
Utilizing the compositionality, our model outperformed the previous SoTA model in an extreme source code summarization task with a considerably smaller number of parameters.
We will further evaluate the capabilities of our model with wider variety of tasks, including natural language processing.

\begin{ack}
We thank Y.~Ikutani for his valuable comments.
This work was supported by JSPS KAKENHI grant number 16H06569, 18K18108, 18K19821, and JP19J20669.
\end{ack}

\bibliographystyle{unsrtabbrvnat}
{\small \bibliography{bib}}

\end{document}